\documentclass[a4paper,english,english,english,english]{article}
 \usepackage{verbatim}

\usepackage[T1]{fontenc}
\usepackage[latin1]{inputenc}
\usepackage{babel}
\usepackage{graphics}
\makeatletter

\usepackage{colacl}
\usepackage{amstex}
\makeatother

\def \stuff {
\parbox{5cm}{
 POS\(_0\)=VBD POS\(_{[-3\ldots-1]}\)=VBZ\\
 POS$_0$=VBD POS$_{-1}$=VBD\\
 POS$_0$=VBD POS\(_{\left[-3\ldots-1\right]}\)=VBP
 }
}

\def \mtarg {
  \parbox{1cm}{\centering VBN\\VBN\\VBN}

}

\makeatother
\begin{document}

\title{Multidimensional Transformation-Based Learning}

\author{Radu Florian\( ^{\dagger } \) \and Grace Ngai\( ^{\dagger ,\ddagger } \)\\
\{rflorian,gyn\}@@cs.jhu.edu\\
\vspace*{0.1cm}\\
\begin{tabular}{cc}
\( ^{\dagger } \)\parbox[t]{5cm}{\centering Johns Hopkins University\\Baltimore, MD 21218,~USA}&
\( ^{\ddagger } \)\parbox[t]{4.3cm}{\centering Weniwen Technologies\\ Hong Kong}\\
\end{tabular}}

\maketitle
\begin{abstract}
This paper presents a novel method that allows a machine learning
algorithm following the transformation-based learning paradigm \cite{brill95:tagging}
to be applied to multiple classification tasks by training jointly
and simultaneously on all fields. The motivation for constructing
such a system stems from the observation that many tasks in natural
language processing are naturally composed of multiple subtasks which
need to be resolved simultaneously; also tasks usually learned in
isolation can possibly benefit from being learned in a joint framework,
as the signals for the extra tasks usually constitute inductive bias.

The proposed algorithm is evaluated in two experiments: in one, the
system is used to jointly predict the part-of-speech and text chunks/baseNP
chunks of an English corpus; and in the second it is used to learn
the joint prediction of word segment boundaries and part-of-speech
tagging for Chinese. The results show that the simultaneous learning
of multiple tasks does achieve an improvement in each task upon training
the same tasks sequentially. The part-of-speech tagging result of
96.63\% is state-of-the-art for individual systems on the particular
train/test split.
\end{abstract}

\section{Introduction}

Transformation-based learning (TBL) \cite{brill95:tagging} is one
of the most successful rule-based machine learning algorithms. It
is a flexible and powerful method which is easily extended to various
tasks and domains, and it has been applied to a wide variety of NLP
tasks, including part of speech tagging \cite{brill95:tagging}, parsing
\cite{brill96:transformation_parsing} and phrase chunking \cite{ramshaw94,florian00:tbldt}.
It achieves state-of-the-art performance on several tasks, and has
been shown to be fairly resistant to overtraining \cite{ramshaw94}.

The processing of natural language text is usually done through a
pipeline of well defined tasks, each extracting specific information.
For instance, one possible sequence of actions performed could be: 

\begin{enumerate}
\item Tokenize the text;
\item Associate part-of-speech tags with each word;
\item Parse each sentence;
\item Identify and label named entities;
\item Resolve anaphora.
\end{enumerate}
In the above scenario, each task is well-defined in itself and is
often performed independently and in a specific order. There are NLP
tasks, however, which consist of closely-related sub-tasks, where
the order and independence is hard to determine \,---\, for example,
the task of part-of-speech (POS) tagging in highly inflective languages
such as Czech. A POS tag in Czech consists of several sub-tags, including
the main part-of-speech (e.g. noun, verb), a detailed part-of-speech
(e.g. past tense verb, genitive noun, etc), gender, case, number and
some other 11 sub-tags. Allowing a system to learn the sub-tasks jointly
is beneficial in this case, as it eliminates the need to define a
learning order, and it allows the true dependencies between the sub-tasks
to be modeled, while not imposing artificial dependencies among them.

The multi-task classification approach we are presenting in this paper
is very similar to the one proposed by \newcite{caruana97multitask}.
Instead of using neural network learning, we are modifying the transformation-based
learning to able to perform multiple-task classification. The new
framework is tested by performing joint POS tagging and base noun-phrase
(baseNP) chunking on the Penn Treebank Wall Street Journal corpus
\cite{penntreebank}, and simultaneous word segmentation and POS tagging
on the Chinese Treebank's \cite{xia00}. In both experiments the jointly
trained system outperforms the sequentially-trained system combination. 

The remainder of the paper is organized as follows: Section 2 briefly
presents previous approaches to multi-task classification; Section
3 describes the general TBL framework and the proposed modifications
to it; Section 4 describes the experiments and the results; Section
5 does a qualitative analysis of the behavior of the system, and Section
6 concludes the paper.

\section{Previous Work}

\subsubsection*{Multitask Learning}

\cite{caruana97multitask} analyzes in depth the multi-task learning
(MTL) paradigm, where individual related tasks are trained together
by sharing a common representation of knowledge, and shows that such
a strategy obtains better results than a single-task learning strategy.
The algorithm of choice there is a backpropagation neural network,
and the paradigm is tested on several machine learning tasks, including
1D-ALVINN (road-following problem), 1D-DOORS (locate doorknobs to
recognize door types) and pneumonia prediction. It also contains an
excellent discussion on how and why the MTL paradigm is superior to
single-task learning.

\subsubsection*{Combining Classifications}

A straightforward way of addressing a multiple classification problem
is to create a new label for each observed \textit{combination} of
the original sub-tags; \newcite{hajic98:PDT_tagSet} describes such
an approach for performing POS tagging in Czech. While it has the
advantage of not modifying the structure of the original algorithm,
it does have some drawbacks:

\begin{itemize}
\item By increasing the range of possible classifications, each individual
tag label will have fewer samples associated with it, resulting in
data sparseness. For example, in Czech, {}``glueing{}'' together
the subtags results in 1291 part-of-speech tags, a considerably larger
number than in number of POS tags in English \,---\, 55. It is arguable
that one would need one order or magnitude more Czech data than English
data to estimate similarly well the same model parameters. 
\item No new class labels will be generated, even if it should be possible
to assign a label consisting of sub-parts that were observed in the
training set, but whose combination wasn't actually seen. 
\end{itemize}

\subsubsection*{N-Best Rescoring}

Another trend in a 2-task classification is to use a single-task classifier
for the first task to output \emph{n-best} lists and then use a classifier
trained on the joint tasks to select the best candidate that maximizes
the joint likelihood. \newcite{xun00:basenp} performs a joint POS
tagging / baseNP classification by using a statistical POS tagger
to generate n-best lists of POS tags, and then a Viterbi algorithm
to determine the best candidate that maximizes the joint probability
of POS tag/ baseNP chunk. \newcite{chang93:chinese_wordseg_pos} uses
a similar technique to perform word-segmentation and POS tagging in
Chinese texts. In both approaches, the joint search obtains better
results than the ones obtained when the search was performed independently.

\section{Multi-task Training with Transformation-Based Learning}

The multi-dimensional training method presented in this paper learns
multiple \emph{related} tasks in parallel, by using the domain specific
signals present in each training stream. The tasks share a common
representation, and rules are allowed to correct any of the errors
present in the streams, without imposing ordering restrictions on
the type of the individual errors (i.e. learn POS tagging before baseNP
chunking).

Transformation-based learning (TBL) is well-suited to perform in such
a framework:

\begin{itemize}
\item Partial classifications are easily accommodated in the TBL paradigm,
as features of the samples (e.g. word, gender, number);
\item The system can learn rules that correct one sub-classification, then
use the corrected sub-classification to correct the other classifications,
in a seemingly interspersed fashion, as dictated by the data;
\item The objective function used in TBL usually is the evaluation measure
of the task (e.g. number of errors, F-Measure). This allows the algorithm
to work directly toward optimizing its evaluation function.\footnote{%
This distinguishes TBL as a error-driven approach from other feature-based
methods such as maximum entropy which adjust parameters to maximize
the likelihood of the data; the latter may not be perfectly correlated
with the classifier performance.
}

\end{itemize}

\subsection{The Standard TBL Algorithm}

The central idea behind transformation-based learning is to induce
an ordered list of rules which progressively improve upon the current
state of the training set. An initial assignment is made based on
simple statistics, and rules are then greedily learned to correct
the existing mistakes until no net improvement can be made.

The use of a transformation-based system assumes the existence of
the following:

\begin{itemize}
\item An initial state generator; 
\item A set of allowable transform types, or templates; 
\item An objective function for learning \,---\, typically the evaluation
function. 
\end{itemize}
Before learning begins, the training corpus is passed through the
initial state generator which assigns to each instance some initial
classification. The learner then iteratively learns an ordered sequence
of rules:

\begin{enumerate}
\item \label{tbl:begin} For each possible transformation, or rule, \( r \)
that can be applied to the corpus:

\begin{enumerate}
\item Apply the rule to a copy of the current state of the corpus, 
\item Score the resulting corpus with the objective function; compute the
score associated with the applied rule (\( f\left( r\right)  \))
(usually the difference in performance)
\end{enumerate}
\item If no rule with a positive objective function score exists, learning
is finished. Stop. 
\item Select the rule with the best score; append it to the list of learned
rules; 
\item Apply the rule to the current state of the corpus. 
\item Repeat from Step \ref{tbl:begin}. 
\end{enumerate}
At evaluation time, the test data is first initialized by passing
it through the initial state generator. The rules are then applied
to the data in the order in which they were learned. The final state
of the test data after all the rules have been applied constitutes
the output of the system.

\subsection{Multi-task Rule Evaluation Function}

The algorithm for the multidimensional transformation-based learner
(mTBL) can be derived easily from the standard algorithm. The only
change needed is modifying the objective function to take into account
the current state of all the subtags (the classifications of the various
sub-tasks): \[
f\left( r\right) =\sum _{s\textrm{ }{\textrm{sample}}}\sum _{i=1}^{n}w_{i}\cdot \left( S_{i}\left( r\left( s\right) \right) -S_{i}\left( s\right) \right) \]
 where 

\begin{itemize}
\item \( r \) is a rule 
\item \( r\left( s\right)  \) is the result of applying rule \( r \) to
sample \( s \)
\item \( n \) is the number of tasks 
\item \( S_{i}\left( s\right)  \) is the score on sub-classification \( i \)
of sample \( s \) (1 if is correct and \( 0 \) otherwise). 
\item \( w_{i} \) represent weights that can be assigned to tasks to impose
priorities for specific sub-tasks. In the experiments, all the weights
were set to \( 1 \).
\end{itemize}

\section{Experiments\label{section:experiments}}

\begin{table*}
{\centering \begin{tabular}{|c||c||c|c||c|c|}
\hline 
System&
\multicolumn{1}{|c||}{ POS Accuracy}&
\multicolumn{2}{|c||}{BaseNP Chunking}&
\multicolumn{2}{|c|}{Text Chunking}\\
\cline{3-4} 
\cline{3-3} \cline{4-4} \cline{5-5} \cline{6-6} 
\multicolumn{1}{|c||}{}&
\multicolumn{1}{|c||}{}&
Accuracy&
F\( _{\beta =1} \)&
Accuracy&
F\( _{\beta =1} \)\\
\hline
\hline 
{\footnotesize Sequential POS, Base NP}&
96.45 \%&
97.41 \%&
92.49&
-&
-\\
\hline 
{\footnotesize Joint POS / Base NP }&
96.55 \%&
97.65 \%&
92.73&
-&
-\\
\hline 
{\footnotesize Sequential POS, Text Chunking}&
 96.45 \%&
96.97 \%&
92.92&
95.45 \%&
 92.65\\
\hline
{\footnotesize Joint POS / Text Chunking}&
 \textbf{96.63} \%&
97.22 \%&
\textbf{93.29}&
95.82 \%&
 \textbf{93.12} \\
\hline
\end{tabular}\par}

\caption{Part-of-Speech Tagging and Text Chunking\label{pos-chunking}}
\end{table*}
For our experiments, we adapted the fast version of TBL described
in \cite{ngai01} for multidimensional classification. All the systems
compared in the following experiments are TBL-based systems; the difference
we are interested in is the performance of a sequential training of
systems versus the the performance of the system that learns the tasks
jointly.

\subsection{English POS tagging and Base Noun Phrase/Text Chunking\label{English-proc}}

The first experiment performed was to learn to jointly perform POS
tagging and text/baseNP chunking on an English corpus. This section
will give an overview of the task and detail the experimental results.

\subsubsection{Part-of-Speech Tagging}

Part-of-speech (POS) tagging is one of the most basic tasks in natural
language processing. It involves labeling each word in a sentence
with a tag indicating its part-of-speech function (such as noun, verb
or adjective). It is an important precursor to many higher-level NLP
tasks (e.g. parsing, word sense disambiguation, etc).

There has been much research done in POS tagging. Among the more notable
efforts were Brill's transformation-based tagger \cite{brill95:tagging},
Ratnaparkhi's Maximum Entropy tagger \cite{ratnaparkhi96:tagging},
and the TnT tagger \cite{brants00}, which features an ngram approach.
State-of-the-art performance on POS tagging in English for individual
systems is around 96.5\%-96.7\% accuracy on the Penn Treebank Wall
Street Journal corpus.

\subsubsection{Text and Base Noun Phrase Chunking}

Text chunking and base noun phrase (baseNP) chunking are both subproblems
of syntactic parsing. Unlike syntactic parsing, where the goal is
to reconstruct the complete phrasal structure of a sentence, chunking
divides the sentence into non-overlapping, flat phrases. For baseNP
chunking, the identified phrases are the non-recursive noun phrases;
in text chunking, the identified phrases are the basic phrasal structures
in the sentence (e.g. verb phrase, noun phrase, adverbial phrase,
etc) \,---\, words are considered to belong to a chunk given by the
lowest constituent in the parse tree that dominates it.

Even though the identified structures are much less complex than that
in syntactic parsing, text chunks are useful in many situations where
some knowledge of the syntactic relations are useful, as they constitute
a simplified version of shallow parsing.
\begin{table}
{\centering \begin{tabular}{|c|c|c|}
\hline 
Word &
 POS tag &
 Text Chunk Tag\\
\hline
A.P. &
 NNP &
 B-NP\\
 Green &
 NNP &
 I-NP\\
 currently &
 RB &
 B-ADVP\\
 has &
 VBZ &
 B-VP\\
 2,664,098 &
 CD &
 B-NP\\
 shares &
 NNS &
 I-NP\\
 outstanding &
 JJ &
 B-ADJP\\
 . &
 . &
 O  \\
\hline
\end{tabular}\par}

\caption{Example Sentence from the corpus.\label{table:example}}
\end{table}

The established measure in evaluating performance in these tasks is
the F-measure, which is based on precision and recall:\[
F_{\beta }=2\cdot \frac{\beta ^{2}P\cdot R}{\beta ^{2}P+R}\]
where \[
\begin{array}{c}
P=\frac{\#\textrm{ correctly found chunks}}{\#\textrm{ found chunks}}\\
R=\frac{\#\textrm{ correctly found chunks}}{\#\textrm{ true chunks}}
\end{array}\]

Ramshaw \& Marcus \shortcite{ramshaw99:basenp} were the first to consider
baseNP chunking as a classification task; the same approach can be
applied to text chunking. Because the structure is not recursive,
any possible pattern can be described by assigning to each word a
tag corresponding to whether the word starts, is inside or is outside
of a noun phrase ({}``B{}'', {}``I{}'' or {}``O{}''). Ramshaw
\& Marcus trained their system on Sections 15-18 of the Penn Treebank
Wall Street Journal Corpus \cite{penntreebank}, and achieved an F-Measure
performance of 92.0\%. Other notable (and comparable) efforts include
Munoz et al. \shortcite{roth99:basenp}, who used a Winnow-based system
to achieve an F-Measure of 92.7, and Tjong Kim Sang \shortcite{tks00:npCombo}
who used a combination of 4 different systems to achieve an F-Measure
of 93.2.

There has also been interest in text chunking in recent years. Similar
to the base noun phrase task, each word is assigned a tag corresponding
to the lowest constituent in the parse tree that dominates it (e.g.
NP, VP, PP, ADVP). In addition, a chunk tag has a prefix that specifies
whether the word is the first one in its chunk, or somewhere in the
middle ({}``B{}'' and {}``I{}''). Table \ref{table:example} shows
an example sentence with POS and text chunk tags. 

Text chunking was featured as the shared task at CoNLL 2000 \cite{tjong00:conll_shared};
the training set consisted of the sections 15-18 of the Penn Treebank,
and as test the section 20 of the same corpus was selected. The task
attracted several participants; the best individual system achieved
an F-Measure performance of 92.12, and combination systems obtained
up to 93.48 F-measure.

For the sake of facilitating comparisons with previously published
results, this paper will report results on both text and base noun
phrase chunking. Even if the base noun phrase is a subtask of text
chunking, the slightly differing conventions used for pulling the
structures out from the Treebank it is not redundant to present both
experiments.

\subsubsection{Experimental Results}

The corpus used in these experiments is the Penn Treebank Wall Street
Journal corpus. The training data consists of sections 02-21 and the
test data consists of section 00.

As an initial state, each word is assigned the POS tag that it was
most often associated with in the training set and the chunk tag that
was most often associated with the initial POS tag. The rules templates
are allowed to consult the word, POS tag and chunk tag of the current
word and up to three words/POS tags/chunk tags to either side of it.
Some of the rule templates modify the chunk tags and other ones modify
the POS tag.

Table \ref{pos-chunking} presents the results of the 4 experiments.
Interestingly enough, extracting the base NP structures by performing
text chunking obtains better F-measure, but the difference could be
an artifact of the two annotation schemes not agreeing on what constitutes
a noun phrase\footnote{%
We tried, as much as possible, to report numbers that are comparable
with other published numbers, rather than ensuring that the results
are consistent among text and baseNP chunking.
}.

\begin{figure*}
{\centering \resizebox*{0.85\textwidth}{!}{\includegraphics{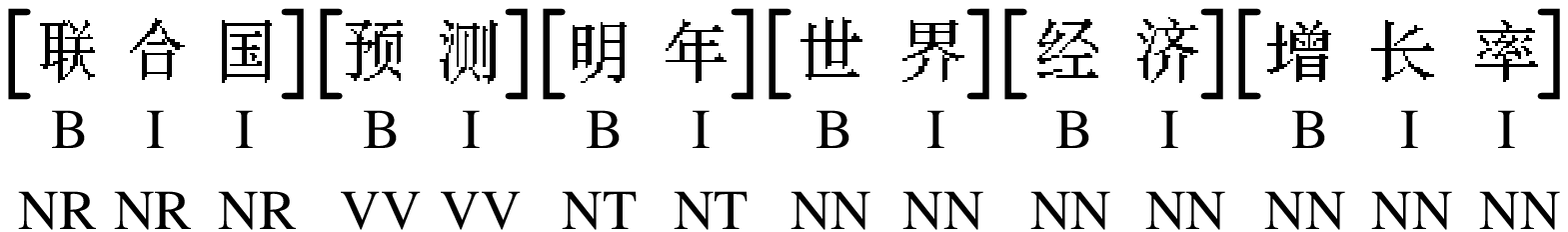}} \par}

\vspace{0pt}
\caption{Chinese Sentence Segmentation and POS Tagging\\(Translation: {}``The
United Nations makes a prediction on next year's world economy growth
rate.{}'')\label{example:chinese}}
\end{figure*}
It can be seen that training the systems jointly results in better
performance, especially for the chunking tasks. An analysis of the
algorithm's behavior is given in Section \ref{analysis}. When trained
jointly on the text chunking task, the POS tagger obtains an accuracy
of 96.63\%, which is among the state-of-the-art results for individual
systems.

\subsection{Chinese Word Segmentation and POS Tagging\label{Chinese-proc}}

\subsubsection{Problem description}

Word segmentation is a problem which is unique to some Asian languages
such as Chinese and Japanese. Unlike most Indo-European languages,
these languages are not written with any spaces or characters which
indicate the boundaries between words. Since most existent NLP techniques
are based on processing words (rather than streams of characters),
word segmentation is a rather necessary task \,---\, it attempts to
word-delimit a text by inserting {}``spaces{}'' between characters
where a pre-defined word boundary exists.

One major difficulty of performing Chinese word segmentation stems
from the ambiguity of the task. The concept of a word is not clearly
defined: experiments involving native speakers show an agreement rate
of only around 75\% \cite{sproat96:chinese_wordseg,wu94:chinese_wordseg}.

Since the character segments (words) obtained from segmentation are
non-overlapping, the task can be viewed as a classification task in
the same way as baseNP chunking can. Each character is tagged with
a tag that marks it as either beginning a word ({}``B{}'') or inside
the word ({}``I{}'').

Once the words in a sentence have been identified, part-of-speech
tags can be assigned to words in the same fashion as in English. The
Chinese Treebank \cite{xia00} assigns a total of 33 POS tags. Figure
\ref{example:chinese} shows an example sentence from the Chinese
Treebank that has been annotated with both word segment and POS tags.

Chinese word segmentation and part-of-speech tagging has been extensively
explored in the literature, and dictionary-based methods can usually
achieve extremely high accuracies on the task. However, the inherent
ambiguity of the problem makes system comparison very difficult. The
usual method of evaluating a segmented word as correct as long as
it is not unacceptably wrong also does not make it easy to objectively
compare performances across systems.

Among the machine learning algorithms which have been applied to Chinese
word segmentation, \newcite{palmer97:chinese_wordseg} and \newcite{hockenmeier98:chinese_wordseg}
used transformation-based learning to tackle the problem. Their approach
was to view an example as being the space between two characters.
The rules learned would then insert, delete and move boundary indicators
to obtain the desired words. Hockenmeier \& Brew achieved an F-Measure
performance of 87.8 after training on a corpus of 100,000 words, and
Palmer's system achieved an F-Measure of 87.7 on a corpus of 60,000
words.\footnote{%
Because of the different corpora used in training and testing, the
results are not directly comparable.
}

\subsubsection{Experimental Results}

\begin{table*}
{\centering {\small\begin{tabular}{|c|c|c|}
\hline 
System&
\parbox{1.2in}{\centering Word Segmentation\\ (F-measure)}&
POS Accuracy\\
\hline 
Joint Text Segmentation/POS tagging&
93.55&
88.86 \%\\
\hline 
Text Segmentation, then POS tagging&
93.48&
88.13 \%\\
\hline
\end{tabular}}\par}

\vspace{0pt}
\caption{Chinese Text Processing: Text Segmentation and POS Tagging\label{chinese-results}}
\end{table*}
The corpus used in our experiment is the Chinese Treebank. 80\% (3363
sentences, 141702 words) were randomly selected as training set and
the remaining 20\% of the sentences (820 sentences, 19392 words) were
held out as test set. Since this corpus was released very recently,
we believe that this are the first published results on the Chinese
Treebank.

In the initial state, each Chinese character was assigned the word
segment and POS tag that it was most often associated with in the
training set. The rule predicates are based on conjunctions on the
information (character, POS tag, segmentation tag) for the 3 characters
on either side of the character to be changed.

To evaluate the performance of our system on word segmentation, the
annotations in the Chinese Treebank were considered to be the gold
standard \,---\, i.e. a segmented word is incorrect if it disagrees
with the Treebank.

Since we are training the system to perform part-of-speech tagging
together with word segmentation, and labeling part-of-speech tags
per character basis, there is a possibility that the system may assign
different labels to individual characters inside the same word. In
such a situation, a word is assigned the part-of-speech tag that was
assigned to the majority of the characters it contains.

Table \ref{chinese-results} presents the results of the experiments.
mTBL achieves a respectable performance on word segmentation, and
when compared to a sequential system of segmenting and then tagging
words, mTBL outperforms the sequential system significantly for POS
tagging. 

\section{Analysis\label{analysis}}

In the previous section, we presented the results of several experiments,
in which training simultaneously on multiple tasks consistently outperformed
than the sequential training on one task at a time. In this section,
we will analyze some of the advantages of the joint system, considering
the POS tagging/text chunking experiment as case study.
\begin{figure*}
{\centering \begin{tabular}{|c|c||c|c|}
\hline 
Condition&
Change To&
Condition&
Change To\\
\hline 
POS\( _{0} \)=VBD Chunk\( _{0} \)=I-VP &
VBN&
\stuff&
\mtarg\\
\hline
\multicolumn{2}{c}{Jointly trained system}&
\multicolumn{2}{c}{POS only}\\
\end{tabular}\par}

\vspace{0pt}
\caption{Example of Learned Rules\label{VBNrules}}
\end{figure*}

Sentence chunks offer the POS tagger a way to generalize the contexts;
this is due to a limitation of the template types. Our choice of predicate
templates is a conjunction of feature identities (e.g. if the previous
POS tag is {}``TO{}'') and/or an atomic predicate that can examine
one of the previous \( k \) words (e.g. if one of the previous 3
tags if {}``MD{}''). Extending the template structure to include
disjunction would create a very highly dimensional search space, making
the problem intractable. However, introducing the sentence chunk tags
can alleviate this problem; in Figure \ref{VBNrules}, we have shown
the rules learned by the 2 systems to resolve the disambiguation between
the POS tags VBD (past tense) and VBN (past participle). Most verbs
in English display the same form while used as past tense or past
participle (all regular ones, plus some of the irregular ones), but
their grammatical use of the 2 forms is completely different: past
tenses usually create predicates by themselves (e.g {}``he drank
water{}''), while the part participles are part of a complex predicate
({}``I have been present{}'', {}''he was cited{}'') or are used
as adjectives (e.g. {}``the used book{}''). Figure \ref{VBNrules}shows
that the jointly trained system can make the distinction using just
one rule, by deciding that if the current verb is inside of a verb
phrase, then the form should be the past participle. The TBL system
that was trained only on the POS task breaks this rule into several
particular cases, which are not learned contiguously (some other rules
were learned in-between). Using more general rules is desirable, as
it will not {}``split{}'' the data as much as a more particular
rule would do. In the end, the jointly trained system made 30 less
POS errors on samples labeled VBN or VBD on which the displayed rules
applied (there are 2026 samples labeled VBN or VBD in the test corpus).

A second reason for the better behavior of the jointly trained system
is that, for the POS case, the systems' performance is approaching
the inter-annotator agreement, and therefore further improvement is
difficult\footnote{%
There are areas in which the systems can be improved; the classification
for words unseen in the training data is one of them.
}. By training the system jointly, the system can choose to model the
problematic cases (the ones that are truly ambiguous or the ones on
which the annotators disagree consistently) in such a way that the
second task is improved.

One advantage the multi-task system has compared with the independently
trained system is the consistency between the quality of data received
at training time and the one received at test time. In the case of
the sequential approach, the POS assignment is much more accurate
during the training process (being the output of the POS system on
the training data - 98.54\%) than during testing (96.45\%), while
for the jointly trained system, the input has the same accuracy during
testing and training\footnote{%
Assuming, of course, that the test and train data have been drawn
from the same distribution.
} . Also, by starting from a less accurate initial point, the joint
system is able to filter out some of the noise, resulting in a more
robust classification. 

To examine this aspect in more detail, let us consider the initial
conditional probability (as assigned in the initialization phase)
of a chunk tag \( c \) given a word \( w \) during testing \( P_{t}\left( c|w\right)  \)
and training \( P_{T}\left( c|w\right)  \). A measure of disagreement
between the probability distributions during training and testing
is the Kullbach-Leibler distance\[
D\left( P_{t}\left( \cdot |w\right) ,P_{T}\left( \cdot |w\right) \right) =\sum _{c}P_{t}\left( c|w\right) \cdot \log \frac{P_{t}\left( c|w\right) }{P_{T}\left( c|w\right) }\]

\begin{figure}
{\centering \hspace*{-7mm}\resizebox*{7cm}{!}{\includegraphics{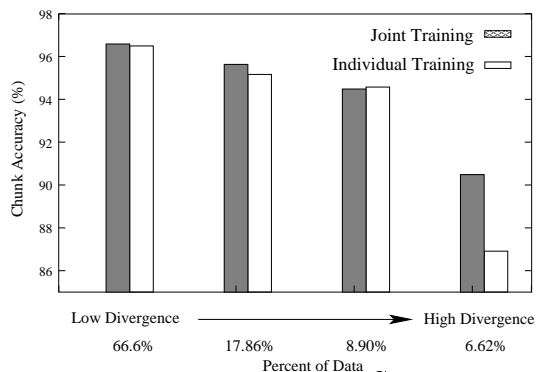}} \vspace*{-5mm}\par}

\vspace{0pt}
\caption{Performance Comparison\label{perplexity-split}}
\end{figure}

Figure \ref{perplexity-split} presents a decomposition of the performance
of the two systems, based on partition of words into 4 classes based
on probability distribution divergence between the initial train and
test data probability distribution. The jointly trained system significantly
outperforms the individually trained chunker on the class with the
highest divergence \,---\, the one that matches the least the training
data, proving that, indeed, the mTBL system is more robust.

\section{Conclusion}

In this paper, we have presented a novel method of using a transformation-based
learner to train on and output multi-task classifications. The simultaneous
multiple classification allows the system to learn from the signals
presented in the training streams, and learned rules can choose to
correct any of the streams, as dictated by the data. Our experiments
show that, in both English part-of-speech tagging/text chunking and
Chinese word segmentation/part-of-speech tagging, the performance
of the jointly trained system outperformed each individually trained
system. In the case of POS tagging for English, the resulting performance
(96.63\%) is very close to state-of-the-art, as reported by \cite{ratnaparkhi96:tagging,brants00};
the difference in performance against the sequential training method
is statistically significant for most tasks (except for Chinese text
segmentation), as verified by a t-test.

Future directions of research include applying the method to part-of-speech
tag inflective languages, and extend the experiment described in Section
\ref{Chinese-proc} by incorporating text chunking. Also, an interesting
research question related to TBL concerns the main design issue in
TBL: way the rule templates are chosen\footnote{%
And not only for TBL \,---\, feature selection is an interesting problem
for other tasks as well. 
}. The authors plan to investigate automated, principled ways to select
the most appropriate rule templates for a given task.

\section{Acknowledgments}

The authors would like to thank David Yarowsky and Ellen Riloff for
valuable advice and suggestions regarding this work, and also thank
the anonymous reviewers for their suggestions and comments. The work
presented here was funded by Grants IIS-9985033 and IRI-9618874, as
well as Weniwen Technologies.

\bibliographystyle{acl}
\bibliography{conll20011}

\end{document}